\documentclass[11pt]{article}
\usepackage{arabtex}
\usepackage{coling2020}
\usepackage{times}
\usepackage{latexsym}
\usepackage{utf8}
\usepackage[utf8]{inputenc}
\setcode{utf8}
\usepackage{graphicx}
\usepackage{multirow}
\usepackage{url}

\emergencystretch=\maxdimen
\colingfinalcopy % Uncomment this line for the final submission

\title{Fake or Real?
A Study of Arabic Satirical Fake News}

\author{Hadeel Saadany and Emad Mohamed \\
  Research Group in Computational Linguistics \\
  University of Wolverhampton, UK \\
  {\tt h.a.saadany@wlv.ac.uk} \\ {\tt E.Mohamed2@wlv.ac.uk} \\\And
  
  Constantin Or\u{a}san \\
  Centre for Translation Studies \\
  University of Surrey, UK \\
  {\tt C.Orasan@surrey.ac.uk} \\}

\date{}

\begin{document}
\maketitle

\begin{abstract}
One very common type of fake news is satire which comes in a form of a news website or an online platform that parodies reputable real news agencies to create a sarcastic version of reality. This type of fake news is often disseminated by individuals on their online platforms as it has a much stronger effect in delivering criticism than through a straightforward message.  However, when the satirical text is disseminated via social media without mention of its source, it can be mistaken for real news. This study conducts several exploratory analyses to identify the linguistic properties of Arabic fake news with satirical content. It shows that although it parodies real news, Arabic satirical news has distinguishing features on the lexico-grammatical level. We exploit these features to build a number of machine learning models capable of identifying satirical fake news with an accuracy of up to 98.6\%. The study introduces a new dataset (3185 articles) scraped from two Arabic satirical news websites (`Al-Hudood' and `Al-Ahram Al-Mexici') which consists of fake news. The real news dataset consists of 3710 articles collected from  three official news sites: the `BBC-Arabic', the `CNN-Arabic' and `Al-Jazeera news'. Both datasets are concerned with political issues related to the Middle East.  

\end{abstract}

\section{Introduction}

% The following footnote without marker is needed for the camera-ready
% version of the paper.
% Comment out the instructions (first text) and uncomment the 8 lines
% under "final paper" for your variant of English.
% 
\blfootnote{
    %
    % for review submission
    %
    \hspace{-0.65cm}  % space normally used by the marker
    % Place licence statement here for the camera-ready version. See
    % Section~\ref{licence} of the instructions for preparing a
    % manuscript.
    %
    % final paper: en-uk version 
    
    \hspace{-0.65cm}  % space normally used by the marker
    This work is licensed under a Creative Commons 
    Attribution 4.0 International Licence.
    Licence details:
    \url{http://creativecommons.org/licenses/by/4.0/}.
    
    % % final paper: en-us version 
    %
    % \hspace{-0.65cm}  % space normally used by the marker
    % This work is licensed under a Creative Commons 
    % Attribution 4.0 International License.
    % License details:
    % \url{http://creativecommons.org/licenses/by/4.0/}.
}

In September 2015, the Egyptian army Apache helicopters blasted a convoy of Mexican tourists in the Sinai peninsula which resulted in the killing of 12 tourists as it mistakenly took their four-wheel vehicles for terrorist targets\footnote{\url{https://www.bbc.co.uk/news/world-middle-east-34241680} (last accessed on 18 Sept 2020)}. Days later, two renowned Egyptian talk-show hosts announced that the Mexican president stated that he completely understands the Egyptian authorities' intention behind the killing of the Mexican tourists as Egypt has the right to defend its land against possible terrorist threats. A hashtag ``\#thankyou\_Henrico\_Iglassios" went viral on twitter by Egyptian pro-government twitter accounts in an attempt to show gratitude to the Mexican president for his understanding. The absurdity of the Mexican president's statement was realised a little too late. It was soon discovered that the Mexican president is not actually called `Henrico Iglassios' and that this statement was never issued by any of the Mexican authorities who were in reality extremely angry at the incident. The source of all this commotion was an article published on an online news website, `Al-Ahram Al-Mexici'\footnote{\url{https://rassd.com/157396.htm} (last accessed on 18 Sept 2020)}. The latter is a satirical fake newspaper that constantly posts a mockery of current events in a style that meticulously imitates the style and content of Egyptian government news agencies.

The Egyptian-Mexican incident is not the first in Egypt or in other parts of the world. There are similar incidents where satirical news is mistakenly propagated by officials or social media users because they either missed the satirical gesture or they did not investigate the source. For instance,  \textit{The Daily Beast}\footnote{\url{https://www.thedailybeast.com/}} compiled a list of nine times the headlines of the satirical online newspaper, \textit{The Onion}\footnote{\url{https://www.theonion.com/}},  fooled famous news outlets into believing their satirical stories. Among those duped were the New York Times, the US Capitol Police, and the Washington Post \cite{dailybeast}. Although satirical news is essentially entertaining, it is potentially deceptive and harmful since the satirical cues could be easily missed specially if the news source is obliterated in a shared social media post or tweet. Moreover, satirical news aims to entertain by crafting unusual or surprising report of current events. According to a NY Times study on the psychology of content sharing, the motivation of almost 50\% of online news sharers is to bring ``entertaining content to others" \cite{brett2012psychology}. This makes the deceptive content of a satirical article more likely to be disseminated online especially when the language is a perfect parody of real news and when the original source is not mentioned.

In this research, we employ inferential statistics to investigate the defining features of Arabic satirical news on the lexico-grammatical level. We also put forward the hypothesis that Arabic satirical fake news can be automatically identified with very high accuracy. We demonstrate the effectiveness of using pre-trained word embeddings as input to statistical and neural models in detecting Arabic fake news articles with a satirical intent. To achieve its objectives, the study presents in Section \ref{sec:literature_review} a brief literature review of satirical fake news detection in general and Arabic satire detection in particular. Section \ref{sec:Data Compiling} explains the steps followed for the compilation of the corpora used in this study and the preprocessing applied to the datasets. Section \ref{sec:Linguistic-Statistical} presents a linguistic-statistical analysis of features specific to Arabic satirical fake news. Section \ref{sec:Classification Models} describes the classification experiments and their evaluation. Section \ref{sec:Error  and Model Analysis} presents an error analysis and a discussion of the classification features. Section \ref{sec:Conclusion} presents a conclusion on the conducted experiments and the linguistic analyses carried out in the study.

\section{Literature Review}
\label{sec:literature_review}

\subsection{Satirical Fake News Detection}
\label{subsec:Satirical Fake News Detection}
`Fake news' is generally defined as any piece of news that is intentionally crafted to present information contradicting the truth to achieve a particular goal \cite{definition_fake}. Research carried out in Natural Language Processing (NLP) and fake news in general differentiates between two types of fake news: a) satire which mimics real news, but still gives cues that it should not be taken seriously, b) intentionally crafted fake information which attempts to convince the reader of its validity such as hoax and propaganda \cite{truthofvaryingshades}. A major difference between satirical fake news detection and other types of fake news is that the research objective is not geared towards truthfulness estimation since the aim is fundamentally to criticise rather than to deceive. Moreover, tracking of meta-data such as news source and analyses of social network activities may not be as crucial in detecting satirical news as is the case with other types of fake news. Research on satirical fake news detection is mainly focused on writing styles, psycho-linguistic features, and atypical lexical choices which trigger a
satirical cue. 

An early attempt to automatically detect satirical fake news has been conducted by Burfoot and Baldwin \shortcite{havingalaugh}. Their contribution lies in constructing a comprehensive set of linguistic features which they present as satire-defining. Although they achieve a high classification accuracy using the standard bag-of-words approach (94.3\%), they improve their accuracy by integrating three major feature sets that they discovered to be peculiar of satirical fake news. These feature sets are: 1) profanity, 2) slang, and 3) semantic invalidity. The semantic invalidity of an article is measured by how likely a named entity is mentioned in an unusual context, which is a common feature of satire \cite{havingalaugh}.

In a more recent study, Rubin et al. \shortcite{fakeortruth} proposes a SVM-based algorithm enriched with satire predicative features to classify satirical and legitimate news articles. They have shown that absurdity in an English satirical news is achieved by a number of linguistic features on the word-level such as use of slang, co-occurrence of named-entities that are not usually related and the presence of unfamiliar named-entities at the end of news articles as an ironic non sequitur \cite{fakeortruth}. The largest data for satirical fake news detection is developed by Yang et al. \shortcite{yang2017satirical} who utilise more than 16,000 satirical news articles collected from 14 satirical websites such as \textit{The Onion} and \textit{The Beaverton}\footnote{\url{https://www.thebeaverton.com/}}. The true news dataset consists of more than 160,000 articles from Google News. They develop a neural network model that analyses a hierarchy of character-word-paragraph-document. To decide which paragraph has a satirical cue, they calculate the satirical degree of a paragraph by comparing it against a learnable satire-aware vector which is based on a psycho-linguistic and stylistic feature set. Their model suggests that English satirical fake news can be predicted automatically with fairly high precision and an accuracy of up to 98.54\% \cite{yang2017satirical}.

A further development on the dataset introduced by Yang et al. \shortcite{yang2017satirical} is provided by De Sarkar et al. \shortcite{de2018attending}. They put forward the claim that pretrained word embeddings alone as input are sufficient in detecting satirical articles and that word-level syntax information only marginally improves classifiers performance. They point out that on the lexical level, the English satirical news article is distinct because of its use of profanity, slang, negative affect, atypical grammar structures and punctuation. They acknowledge that hand-crafted analysis of these features might contribute to a robust classifier, but a neural network model such as a Convolutional Neural Network (CNN) or a Recurrent Neural Network (RNN) trained on word embeddings is sufficient for detecting satirical fake news. Their analysis of the learned models reveals that news satire is decided by a key sentence which usually comes last. Their best classification accuracy is 98.18\% achieved by an LSTM model using English Glove embeddings \cite{de2018attending}.

Rashkin et al. \shortcite{truthofvaryingshades} also uses a large corpus of 14,170 satirical articles from one source, \textit {The Onion} and add a small number of satirical articles (800) from two other sources. Their fake dataset also includes hoax and propaganda articles collected from different sources. Their model is based on hand-crafted analyses of linguistic features on the word-level. They deploy the psycholinguistic lexicon provided by the Linguistic Inquiry and Word Count software (LIWC) to categorise words into groups according to their social function \cite{liwc}. They noticed that one distinctive feature of English satirical news is its prominent use of adverbs whereas true news uses assertive words and is less likely to use hedging words. Their model uses a Max-Entropy classifier achieving an F1 score of 0.65. They achieve a relatively lower accuracy by comparison to other satirical news detection research because their fake news dataset includes both hoax and propaganda \cite{truthofvaryingshades}.

\subsection{Arabic Satire and Fake News Detection}

\label{subsec:Arabic Satire and Fake News Detection}

Despite the fact that fake news is a major problem in the Arab world, a survey of the literature of fake news detection in NLP shows a relatively low number of research dedicated to addressing the problem in the Arabic language as compared to English. Sabbeh and Batwaah \shortcite{arabicnewscredibility} have made an attempt to automatically detect fake Arabic news by proposing a hybrid model based on content and context related features. Their model classifies a small dataset of 800 Arabic political news tweets manually labelled and collected from Twitter. A number of non-linguistic features such as account demographics, account name, number of followers are included as context-related features. They experimented with different machine learning algorithms and their best predictive accuracy was 88.9\% achieved by a J48 Decision tree classifier.

A similar study for automatically measuring the credibility of Arabic web news content and specifically twitter posts is carried out by Al-Eidan et al. \shortcite{measuringcredibilityarabic}. They present a stand-alone application for classifying Arabic tweets according to a three-level credibility measure: credible, not credible and questionable. Their system relies on document level features as well as other relevant meta-linguistic information. The main features used for analysis are: 1) similarity measures between the target document and a verified content, 2) presence or absence of inappropriate words, and 3) presence or absence of \textit{URL} links to credible news sources. Their highest precision and recall for detecting fake news in tweets are 0.76 and 0.79 respectively. They note that their system has the best performance when using the similarity measure as the main feature for classification.  \cite{measuringcredibilityarabic}. 

% A more recent study by Ghanem et al. \shortcite{ghanem2020irony} proposes a multilingual irony detection system which is able to detect irony in Arabic tweets as well as in other languages. They use both feature-based models and neural models with monolingual word embeddings. Their neural architectures (CNNs) achieved an accuracy of 80.5\% for detecting Arabic ironic tweets \cite{ghanem2020irony}. Neural architectures have used extensively in Arabic irony detection tasks (\cite{allaith2019neural,ranasinghe2019rgcl}. 
% A more recent study of Arabic fake news with propagandistic content is carried out by Barr{\'o}n-Cedeno et al.\shortcite{preslavproppy}. They introduce a system (proppy) which indicates the level of propaganda content in a news article. Their system computes a propaganda score using a maximum entropy classifier. Their experimentation shows that writing style, text complexity, character n-grams are more effective than word n-grams in propaganda detection. They observe that character n-grams as features achieve the highest F1 measure (64.45) in classifying propaganda and non-propaganda articles \cite{preslavproppy}.

Although,  Arabic fake news detection research is sparse, there is a relatively large number of satire detection in Arabic NLP research. Arabic satire is studied as one form of irony where the producer is using implicit sentiment to convey a message of criticism. Karoui et al. \shortcite{karoui2017soukhria} present a supervised learning method to capture irony, satire and sarcasm in Arabic tweets. They analyse a small corpus of 5,479 of ironic and non-ironic tweets. They conclude that satire is achieved on the textual level via a number of features that create a surprising effect by breaking one norm or another. Opposite sentiments, indirect reflexive pronouns, idiomatic expressions, false assertions and exaggerated lexicon are some of the lexical features they employ to classify ironic and satirical tweets \cite{karoui2017soukhria}. Their classification model achieves an accuracy of 72.76\% by using all the features at the training stage. 

Another study that examines the distinguishing characteristics of Arabic satire is conducted by Al-Ghadhban et al. \shortcite{arabicsarcasm}. They study satire as a form of sarcasm in Arabic tweets where the user conveys the opposite of what he/she means. The basic assumption behind the study is that frequency of words would be enough in defining satirical and non-satirical tweets. They train their model on a relatively small number of satirical tweets (around 340) and their model achieves an accuracy of 67.6\% \cite{arabicsarcasm}.  As for irony detection, neural architectures have been used extensively in Arabic ironic tweet detection tasks \cite{allaith2019neural,ranasinghe2019rgcl}. A recent study by Ghanem et al. \shortcite{ghanem2020irony} proposes a multilingual irony detection system which is able to detect irony in Arabic tweets as well as in other languages. They use both feature-based models and neural models with monolingual word embeddings. Their neural architectures (CNNs) achieved an accuracy of 80.5\% for detecting Arabic ironic tweets \cite{ghanem2020irony}.

It is important to mention that research in Arabic satire and irony detection mainly focuses on satirical tweets as one form of sarcasm where the author ironically sends a message opposite to his/her intention. Moreover, the analysed corpora are usually relatively small due to the difficulty of extracting tweets that are verifiably satirical. On the other hand, satirical fake news is different in two respects. First, although its objective is to entertain and ridicule, it scrupulously imitates real news register which makes it potentially deceptive. Second, it comes in the form of a news article rather than a short tweet or an online comment. To the best of the authors' knowledge, there is not a study addressing the detection of Arabic satirical news as a potential form of deceptive information. The reason might be the sparsity of the data since there are only very few Arabic news websites that are satirical in nature. The dataset analysed in the present study contributes to the creation of a new corpus for Arabic satire analysis.

\section{Data Compiling and Preprocessing}
\label{sec:Data Compiling}
 The fake news dataset\footnote{The dataset is available at \url{https://github.com/sadanyh/Satirical-Fake-News-Dataset}} used in the present study is a collection of 3185  crawled articles from two different sites which explicitly declare that they publish satirical fake news.\footnote{Beautiful Soup, an open-source Python Library, was used in compiling the fake dataset.}  The first site is `Al-Hudood'\footnote{\url{https://alhudood.net}} (The Limits) which is a satirical online newspaper launched in 2013 by a 34-year-old Jordanian-Palestinian. It is characterised by deadpan headlines reminiscent of the US publication \textit{The Onion}. The site averages 1 million unique visitors a month  \cite{al-hudood}. The second source, `Al-Ahram Al-Mexici'\footnote{\url{http://alahraam.com/}} (The Mexican Ahram) website which is a satirical parody of the main national Egyptian newspaper `Al-Ahram' \cite{ahram}. The two satirical online newspapers mock political leaders and Middle Eastern political events.

The real news dataset comprises 3710 articles from the BBC-News Arabic and CNN-Arabic news datasets which are available on the Sourceforge part of the Arabic Computational Linguistics project \footnote{\url{https://sourceforge.net/projects/ar-text-mining/files/Arabic-Corpora/}} as well as the Al-Jazeera News open-source dataset used in El Kourdi et al. \shortcite{jazeera}. To ensure that  the fake and real datasets talk about similar themes, we chose the articles that belong to the politics section in the three real news corpora. A number of preprocessing operations were conducted on the datasets.

By surveying the corpora, it was observed that there are  phrases typically repeated at the header and footer of each article and in some instances at the middle. For example, ``BBC Arabic", ``\< الشرق الأوسط >" (Middle East), ``\<شارك برأيك >" (Give your opinion) are typical of the BBC News articles. The fake news dataset header and footer repeatedly mentioned phrases such as ``\< الحدود>'' (Limits), ``\< المذكور أعلاه>'' (Above Mentioned), ``\< خاص للحدود> '' (Special for The Limits), and so on. In order to capture these stylistic guidelines, each dataset was merged into one file and frequency dictionaries of unigrams, bigrams and trigrams were created for each file. The top most 10\% of the frequency dictionaries of each dataset included most of the website-specific phrases. Two lists of website-specific phrases indicative of the document source were manually created for each dataset and items in the lists were replaced by a space in the corpus. The cleaning of the datasets also included deletion of formatting quirks specific of each source, non-informative textual features such as special characters, English characters and diacritics. Moreover, Arabic is a quite morphologically rich language where a stem word can be packed with affixes that realise different grammatical functions. Hence segmentation of words into their constituent tokens is  a very common practice in Arabic NLP research. Accordingly, the ArabicSOS, an Arabic language segmenter was employed  to segment both datasets \cite{ArabicSOS}. The classification task was performed on segmented and non-segmented versions of the two datasets.

\section{Linguistic-Statistical Analysis}
\label{sec:Linguistic-Statistical}
After cleaning the data, a preliminary review of the corpus was conducted to detect category-defining features of the fake dataset. Then inferential statistic tests were performed to decide the degree of significance each of these features may have on the classification task.

\subsection{Journalistic Register Measure}
\label{subsec:Journalistic}
According to M.A.K Halliday's Systemic Functional theory of language, \textbf{register} relates to various assumptions about language in a situation type which implicates the realisation of certain selections on the  lexico-grammatical stratum \cite{halliday1988ineffability}. For journalistic register, the expected set of lexico-grammatical features are mainly phrases reflecting the level of news veracity such as the speaker's status, the type of news source, the venue and time of the news and so on. In Arabic journalism, there are typical phrases which are commonly used in spoken and written news publications (e.g. ``\< تقرير أعده>" (Reported by...), ``\< قال الناطق باسم  > " (Stated the spokesman of...), ``\< من جانبه أكد >'' (From his side, he asserted that...). Hence, a native speaker of Arabic would intuitively realise from the first few lines of a text that it is a news article if phrases typical of the journalistic register are present. 

By surveying the fake news dataset, we observed that a satirical news article includes a relatively large number of journalistic cliches. A statistical analysis was, therefore, conducted to explore whether the proliferation of journalistic cliches is a defining characteristic of satirical fake news. First, a hand-crafted list of journalistic cliches mentioned in sample articles of the fake  dataset was created, denoted by $\mathcal{C}$. Then, the multi-set of tokens of a news article is represented by $\Omega$. The journalistic register of a news article was measured as follows\footnote{Adopted from \cite{havingalaugh} and adapted to our task}:
\begin{equation}
    J = \frac{1}{|\Omega|}\sum_{\omega \in \Omega} I_{\mathcal{C}}(\omega)
\end{equation}
where $|\Omega|$ denotes the cardinality of the multiset $\Omega$, and $I_{\mathcal{C}}$ is the indicator function over the set $\mathcal{C}$ defined by:
\begin{equation}    \label{eq:Indicator_Function}
    I_{\mathcal{C}}(\omega) = \left\{
        \begin{array}{ll}
            1 & \quad \omega \in \mathcal{C} \\
            0 & \quad \omega \notin \mathcal{C}
        \end{array}
    \right.
\end{equation}

\begin{figure}[ht]
    \centering
    \includegraphics[scale=0.50,trim={0.5cm .5cm .3cm 0cm},clip]{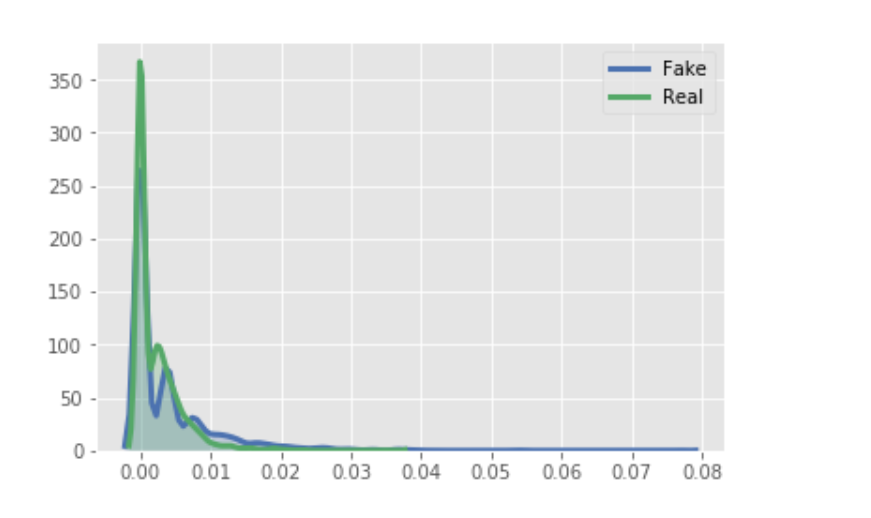}
    \caption{Probability Density Function of Journalistic Register Measure}
    \label{cliche}
\end{figure}

\noindent The function $I_{\mathcal{C}}(\omega)$ takes value 1 if a cliche $\mathcal{\omega}$ is present in an article and 0 if it is not. The journalistic register $J$ is the aggregate number of occurrences of the journalistic cliches normalised to the article size.
A two-tailed T-test with a p-value of .05 was conducted on the average means of journalistic cliches in articles of both datasets. The results were a test statistic of around 0.0628 and a p-value of 0.949, which suggests that there is no statistically significant difference between the use of these phrases in the two news categories (see Figure \ref{cliche} for a probability density function plot of the journalistic register measure in both the fake and real datasets). This points to the
fact that Arabic satirical news meticulously parodies
real journalism register, making it difficult to identify just by using a simple typical news keyword-based approach.

\subsection{Sentiment Intensity Measure}
\label{subsec:Sentiment Intensity}
Professional journalists tend to express opinion objectively to avoid implied bias. By contrast, studies observed that English satirical news includes highly sentimental language for entertainment purpose \cite{golbeck2018fake,yang2017satirical,fakeortruth}. Flouting the journalistic norm by intense affect phrases may well be considered a violation of a conversational maxim which creates satire as observed by philosophers such as Grice \cite{grice1991studies}. Similarly, it was observed that satire in the fake news dataset is often created by an ``emotional imbalance" where high positive and high negative words are present in a context of neutral journalism. To capture the above observation, the degree of sentiment intensity was measured per article against an Arabic emotion lexicon of highly positive and highly negative phrases. We used open-source emotional lexicons for Arabic language \cite{saif3,saif2,saif1}. In an attempt to make the emotional lexicon as exhaustive as possible, a list of hand-picked highly positive and highly negative lexical items, including slang, were also added to the lexicon. The created emotion lexicon is denoted by $\mathcal{L}$. Similar to the Journalistic Register measure, sentiment score was calculated for each article as follows:
\begin{equation}
    S = \frac{1}{|\Omega|}\sum_{\omega \in \Omega} I_{\mathcal{L}}(\omega)
\end{equation}
where $I_{\mathcal{L}}$ is the indicator function over the set $\mathcal{L}$ defined similar to (\ref{eq:Indicator_Function}).

Sentiment intensity measures proved to be statistically significant (statistic $\approx$ 3.27, p-value $\approx$ 0.001) by a T-test with nan-policy=`omit' option to discard all nan values in the measured arrays (the algorithm turned a nan value for words in the sentiment lexicon that did not appear in the dataset). If viewed against the journalistic register measure, it can be concluded that it is with intense sentimentality that satirical and real news diverge whereas with journalistic register they scrupulously converge. 

\subsection{Subjectivity Measure}
\label{subsec:Subjectivity Measure}
Observation of the data has also suggested that fake news articles tend to be written in a more subjective tone. In Arabic, the writer can stylistically identify with the reader by using first person plural (e.g ``\< نري >'' (``we think'')) rather than the first person singular (e.g ``\<أري >'' (``I think'')). On the morphological level, first person plural is realised by the prefix ``\<ن >'' (na) or the suffix ``\< نا>" (naa) which is attached to the verb. Generally speaking, news reporters refrain from using first personal pronouns to create an objective tone, unless it is included in quoted speech. In Arabic satirical news, on the other hand, it was noticed that often words such as ``\<نروي > '' (``we mention'' ), ``\< شارفنا>'' (``we are about to''), and  ``\< نأسف >" (``we regret'' ) are frequently used to report news. The satirist subjectively unites with the reader against the entity that is being targeted for criticism. 

In order to explore whether this type of subjectivity is a distinctive feature of Arabic satirical fake news, the Stanford Log-linear Part-Of-Speech Tagger was used to assign parts of speech tags to tokens of the two datasets \cite{stanford}. The number of tokens that has a POS tag verb and starts or ends with the first person plural inflection was calculated for each dataset and normalised by the number of tokens with `VERB' POS-tag for each dataset. The ratio of first person plural verbs in a satirical fake article to a real news article was 2.2 to 1. Thus, in a satirical news article, it is expected to find double the number of first person plural verbs than in a real one.

\vspace{15pt}

\noindent From the linguistic-statistical analysis, it was concluded that Arabic satirical fake news is distinguished on the lexico-grammatical level. Accordingly, representations of words as features were expected to detect the positive dataset in the classification task. In light of this, in the next section we explore different word vectorisation methods for training both statistical and neural classification models to automatically detect Arabic satirical fake news. 

\section{Classification Models}
\label{sec:Classification Models}
We used a supervised classification approach for the fake news detection task. We developed three types of models: baseline models which rely on statistical machine learning models (Section \ref{subsec:Baseline Models}), classic machine learning models that achieved promising performance in other text classification tasks (Section \ref{subsec:XGBoost}) and neural models that achieved state-of-the-art results in satirical news detection (Section \ref{subsec:CNNs}). We evaluated the models' performance on a held-out test set by measuring the macro average of the Precision, Recall and F1 score of the two categories (\textit{Real} and \textit{Fake}) as well as the total percent accuracy of each model.

\subsection{Baseline Models}
\label{subsec:Baseline Models}
We used a Naive Bayes Multinomial NB classifier as a baseline model with both Bag-Of-Words (BOW) approach and TF-IDF values as numerical features. The CountVectorizer and Tfidf-Vectorizer classes from the Scikit-learn library were used in the experiment \cite{scikit-learn}. The max\_features parameter was set to 1500 to use the most frequently occurring words as features and the max\_df feature value was set to 0.7 to include words that occur in a maximum of 70\% of all the documents. The classifiers were run on tokenised and segmented versions of the fake and real datasets. Additionally, the classifier with BOW values was run on n-gram vectors of a range (2,3) words and character n-grams also with range (2,3) characters. After running the experiments, it was observed that the model with Count Vectors values trained on word-level performed best with an accuracy of 96.23\%. The full evaluation results are presented in Table \ref{acc}.

\subsection{XGBoost with TF-IDF Values}
\label{subsec:XGBoost}
XGBoost is a version of gradient boosted decision tree classifiers. It was the winner algorithm in several classification competitions  such as the KDD cup 2016 competition for predicting research relevance \cite{xg1,xg2}. It has also proved to achieve high performance with an execution speed relatively faster than other classification models \cite{xg3}. We experimented with the XGBoost classifier using the best baseline setting. Accordingly, the XGBoost scikit-learn API was trained on the Count Vector values of article tokens in the fake and real datasets. We used the gbtree booster with a tree maximum depth of 3 to avoid over fitting. The boosting learning rate was set to 0.1 with a binary logistic objective function. With these parameters, the XGBoost classifier surpassed the best performing baseline accuracy by around 0.6\%, achieving an accuracy of 96.81\% (see Table \ref{acc}).

% \subsection{FastText Text Classifier}
% For different text classification tasks, FastText classifier shows results that are on par with deep learning models in terms of accuracy but with a relatively faster performance rate \cite{fasttext}. Accordingly, we experimented with the FastText classifier using Arabic pre-trained word embeddings as input features. We used word embeddings that were trained on an Arabic news corpus \cite{altowayan2016word}. We fine tuned the parameters so that the learning rate was set to 1.0, word bigrams used as input, minimum number of word occurrence was set to 1 and trained on 10 epochs with a hierarchical softmax loss function. With fine-tuned parameters, the FastText achieved an accuracy score of 97.2\%. 

\subsection{Convolutional Neural Networks (CNNs)}
\label{subsec:CNNs}
The last experiment was conducted using a Convolutional Neural Network Architecture (CNN) which is a deep learning classification model. A CNN was chosen for two reasons: first, it has obtained  state-of-the-art results on many text classification tasks in general and satirical fake news detection in particular \cite{zhang2015character,kim2014convolutional,collobert2011natural,de2018attending}. Second, CNNs are effective in extracting morphological information and lexical patterns, both of which have been observed as characteristic features of the positive class in the current study.
The input layer for the CNN was pretrained fastText word-embeddings trained on Arabic Wikipedia \cite{wordembeddings}. The pre-trained word-embeddings covered around 92\% of the corpus vocabulary. The Keras Library sequential models were employed for the task \cite{keras}. A Keras Conv1D layer was added between the Embedding layer of the pre-trained vectors and a GlobalMaxPool1D layer. The filter size for the Conv1D layer was set to 126 filters and a kernel size of 5. Training was done with Adam optimizer, ReLU and sigmoid activation functions, and a binary cross-entropy loss function. The model was initialised by 300-dimensional word-embeddings from the pretrained model and run on 10 epochs of the training data with a batch size of 10 samples for each epoch. CNN outperformed all models, resulting in an accuracy of 98.59\% on the held-out test set. A summary of the evaluation report of the models is given in Table \ref{acc}.

\begin{table}[]
\centering
\begin{tabular}{|l|l|l|l|l|}
\hline
\multicolumn{1}{|c|}{\multirow{2}{*}{\textbf{Model}}} & \multicolumn{4}{c|}{\textbf{Test Set}} \\ \cline{2-5} 
\multicolumn{1}{|c|}{} & \multicolumn{1}{c|}{\textbf{Acc}} & \textbf{P} & \textbf{R} & \textbf{F} \\ \hline
Multinomial NB (Count Vectors) & 96.23 & 96.47 & 96.23 & 96.24 \\ \hline
Multinomial NB (Word-Level   TF-IDF) & 94.63 & 94.71 & 94.64  & 94.64 \\ \hline
Multinomial NB (N-Grams) & 86.08 & 88.03 & 86.09 & 86.22  \\ \hline
Multinomial NB (CharLevel) & 91.73 & 92.00 & 91.74 & 91.75  \\ \hline
Multinomial NB (Count Vectors   segmented) & 95.79 & 95.97 & 95.80  & 95.81 \\ \hline
Multinomial NB (Word-Level   TF-IDF segmented) & 94.49 & 94.65 & 94.49  & 94.51 \\ \hline
XGBoost   with Count Vectors & 96.81 & 96.81 & 96.81 & 96.81      \\ \hline
CNN with   pre-trained word embeddings & \textbf{98.59} & \textbf{98.49} & \textbf{98.61} & \textbf{98.49} \\ \hline
\end{tabular}
\caption{Summary of Accuracy (Acc), Precision (P), Recall (R) and F1 score (F) of classification models} 
\label{acc}
\end{table}

\subsection{Notes on Classification Models}
\label{subsec:Notes on Classification Models}
The results of the classification experiments show that, similar to most research studies of English satirical fake news, the Arabic satirical news can be detected with high accuracy even by the baseline models. It is worth noting that although the neural model (CNN) achieved the highest accuracy, other classifiers such as the XGBoost classifier recorded a very close accuracy figure, but with a much faster execution speed.  Moreover, the precision of all the classification models was relatively higher than the recall as the number of false positive instances (i.e. a real article predicted as fake) was generally low. The CNN model with pre-trained word embeddings, however, was able to detect the fake articles that were predicted as real by other models; recording  the best recall (98.61) and precision (98.49).

It can be claimed that due to its stylistic properties, lexical features per se are sufficient for accurate classification of Arabic satirical fake news. This may not be the case with other types of fake news such as propaganda and hoax where meta-linguistic data can be crucial for falsehood detection. Accordingly, even when a satirical news is disseminated in the form of an online post or a tweet without mention of its source, AI solutions can help in pointing out its deceptive content.

\section{Error Evaluation and Feature Analysis }
\label{sec:Error  and Model Analysis}
This section presents an error analysis of the false negative instances where satirical news was mistaken for real. They constituted the higher number of classification errors with all the models.  We also explore the most informative features for the classification task. Both analyses provided useful insights into the problem of detecting Arabic satirical fake news.

% (see for example
% the confusion matrix of the Count Vector word-level baseline model in figure \ref{cm}).
% \begin{figure}[ht]
%     \centering
%     \includegraphics[scale=.5,trim={0.5cm .5cm .3cm .7cm},clip]{confusion_cnn.PNG}
%     \caption{Confusion Matrix for a Word-level Baseline Model }
%     \label{cm}
% \end{figure}

\subsection{Document-level Error Analysis}
\label{subsec:Document-level}
In order to have a better understanding of the classification results, we analysed some of the fake articles that were labelled ``real''. We noticed that the most common lexical feature of the false negative instances are the proliferation of journalistic cliches (up to 19 phrases per article), which create an almost perfect parody of a real news article. The second observation is that the misclassified articles are all written in Modern Standard Arabic (MSA) and do not contain any slang items. However, there were MSA phrases carrying exaggerated sentiment import that are not common in journalistic register. For example, in one misclassified article, the leader of the Hezbullah Shia political party is described as the ``all-time comrade" (\<  رفيق دربه >) and the ``resistance armour" (\<سلاحه في المقاومه  >) of the leader of the Houthi Movement (a Shia Yemeni Movement involved in the current Yemeni war). Most of the high sentiment lexical items in the misclassified articles were not captured by the emotion lexicon used to measure sentiment intensity. They were mostly coined hyperboles realised on the lexico-grammatical level by common nouns or descriptive adjectives that were crafted by the satirist. 

\subsection{Word-level Error Analysis}
\label{subsec:Word-level}
An attempt was made to find the most effective vectors for the baseline classification models, the CountVectorizer and TF-IDF. This was achieved by zipping the baseline classifier coefficients with the feature names. Accordingly, the log of the estimated probability of a feature given the positive class (fake news) was extracted for the top 30 `most fake' and top 30 `most real' words in the two datasets.

 A number of observations were made on the basis of the analysis of the most-informative feature list. First, the most prominent feature name for fake news is ``\<  المصدر>" (the source). The veracity of the news source is overemphasised in fake news in order to highlight the reliability of the news story and maximise the satirical effect. Similarly, journalistic cliches such as ``asserted" and ``the spokesman" always come up as features of the fake class. Again, the objective is to envelop satire with an aura of credibility. Along with credibility markers, high sentiment polarity words are among the most prominent features (e.g. ``\< عفيفة >" (chaste), ``\<أحمق > "  (foolish), ``\<أجود   > " (most generous)). Moreover, names of a number of  political figures such as Trump and Al-sisi (president of Egypt) are at the top of the fake features. This is quite logical since divisive political figures ignite a rich material for satire. 

As for the real news dataset, the most informative features are  mainly phrases denoting time and venue of the news (e.g. ``\< الاثنين >" (Monday), ``\<الماضي   >" (last), ``\< غزة  >" (Gaza)). Documenting the news report is of higher importance in real news articles. Also, contrary to fake news, with different classification runs, the most informative features of the real news dataset did not once contain an emotive phrase.

\section{Conclusion}
\label{sec:Conclusion}

This paper proposed a novel classification problem for Arabic computational linguistics and machine learning: detecting whether an Arabic news article is true or satirical. The study experimented with different classification techniques including state-of-the-art deep learning models. Experimental results showed a superior performance of a proposed architecture employing the combination of a CNN with pretrained word embeddings. A linguistic-statistical analysis of the satirical dataset gave a deeper insight into the phenomenon of Arabic satirical news. However, the high accuracy of the CNN neural architecture suggests that pre-trained word embeddings as input, without the aid of any syntactic information or any hand-crafted linguistic features, suffices in capturing Arabic satire.  The reason is that a satirical cue is always there in a fake satirical article. Although the writer's aim is for the absurdity of the news to be ultimately realised, the satirical message can still be missed and taken for real.
It is worth mentioning, that the classifier successfully detected the deceptive content of the satirical fake news article published by Al-ahram Al-Mexici which was the cause of the diplomatic crisis between Egypt and Mexico mentioned in the introduction of this paper. The study gives promising evidence that the dissemination of this type of deceptive content via the internet can indeed be automatically prevented by AI solutions.

\bibliographystyle{coling}
\bibliography{coling2020}

\end{document}